\documentclass[10pt,twocolumn,letterpaper]{article}

\usepackage{wacv}              

\usepackage{graphicx}
\usepackage{amsmath}
\usepackage{amssymb}
\usepackage{booktabs}

\usepackage{lipsum}
\usepackage{multirow}
\usepackage{xcolor}

\usepackage{colortbl}
\definecolor{mygray}{gray}{.9}

\usepackage[pagebackref,breaklinks,colorlinks]{hyperref}

\usepackage[capitalize]{cleveref}
\crefname{section}{Sec.}{Secs.}
\Crefname{section}{Section}{Sections}
\Crefname{table}{Table}{Tables}
\crefname{table}{Tab.}{Tabs.}

\def\wacvPaperID{2356} 
\def\confName{WACV}
\def\confYear{2024}

\begin{document}
	
	\title{Prototypical Contrastive Learning-based CLIP Fine-tuning\\ for Object Re-identification}
	
	\author{Jiachen Li and Xiaojin Gong\thanks{The corresponding author.}\\
		College of Information Science and Electronic Engineering, Zhejiang University, China\\
		{\tt\small lijiachen\_isee@zju.edu.cn, gongxj@zju.edu.cn}
}
\maketitle

\begin{abstract}
	This work aims to adapt large-scale pre-trained vision-language models, such as contrastive language-image pre-training (CLIP), to enhance the performance of object re-identification (Re-ID) across various supervision settings. Although prompt learning has enabled a recent work named CLIP-ReID to achieve promising performance, the underlying mechanisms and the necessity of prompt learning remain unclear due to the absence of semantic labels in Re-ID tasks. In this work, we first analyze the role prompt learning in CLIP-ReID and identify its limitations. Based on our investigations, we propose a simple yet effective approach to adapt CLIP for supervised object Re-ID. Our approach directly fine-tunes the image encoder of CLIP using a prototypical contrastive learning (PCL) loss, eliminating the need for prompt learning. Experimental results on both person and vehicle Re-ID datasets demonstrate the competitiveness of our method compared to CLIP-ReID. Furthermore, we extend our PCL-based CLIP fine-tuning approach to unsupervised scenarios, where we achieve state-of-the-art performance. Code is available at \href{https://github.com/RikoLi/PCL-CLIP}{https://github.com/RikoLi/PCL-CLIP}.
\end{abstract}

\section{Introduction}
\label{sec:intro}
Recently, contrastive language-image pre-training (CLIP)~\cite{clip} and other pre-trained vision-language models~\cite{Jia2021,Du2022} have attracted great attention in vision community. CLIP is trained on a dataset of 400 million text-image pairs collected from the internet. Through unsupervised cross-modal contrastive learning from the large-scale dataset, it is capable of learning diverse visual and language semantic concepts and acquiring remarkable transfer abilities. As a result, CLIP has been successfully adapted to various downstream vision tasks, including zero-shot/few-shot recognition~\cite{coop,cocoop}, object detection~\cite{Zhong2022RegionCLIP,Wu2023CORA}, semantic segmentation~\cite{Luo2023SegCLIP}, and more.

\begin{figure}[t]
	\centering
	\includegraphics[width=\linewidth]{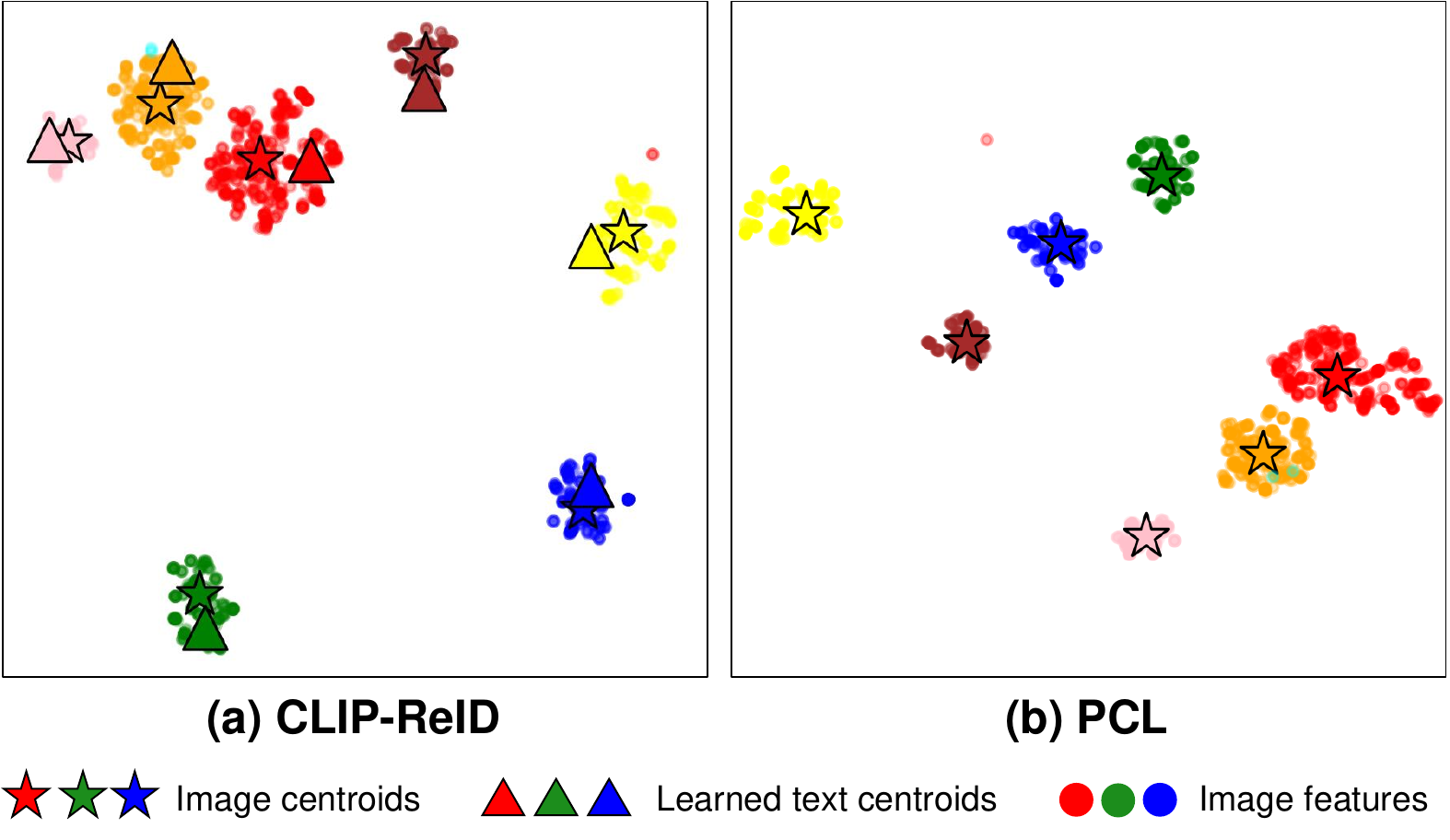}
	\caption{t-SNE~\cite{tsne} visualization of randomly selected 7 IDs from MSMT17. (a) shows that the text centroids learned in CLIP-ReID stage-1 are pretty close to the image centroids, which reveals their implicit equivalence. (b) shows that PCL is also able to learn high-quality feature space only with image centroids. Best view with color.}
	\label{fig:index}
	\vspace{-15pt}
\end{figure}

Various techniques, such as learning additional adaptation layers~\cite{gao2021clip} and learning textual or visual prompts~\cite{coop,cocoop,Jia2022,Khattak2023}, have been developed to facilitate the adaptation of CLIP to specific downstream tasks. Among these adaptation techniques, prompt learning has gained popularity due to its superior performance and low computational cost. Both textual~\cite{coop,cocoop} and multi-modal~\cite{Khattak2023} prompt learning use class names from ground-truth labels to form input text descriptions for recognition or classification tasks. The inclusion of fixed class names helps the CLIP model transfer textual semantics to visual concepts, resulting in great robustness to noise~\cite{Wu2023} and remarkable generalization ability to unseen data~\cite{coop}. 

Unfortunately, in person/vehicle re-identification, class names do not exist as the class labels represent ID indexes and lack semantic meaning. This poses a challenge when employing prompt learning techniques to adapt pre-trained vision-language models. A recent work called CLIP-ReID~\cite{clipreid} addresses this issue by introducing a two-stage strategy. In the first stage, it learns a set of textual prompts for each ID while keeping the CLIP model fixed. Then, in the second stage, the learned prompts are utilized to fine-tune the image encoder of CLIP using a proposed image-to-text cross-entropy loss, in conjunction with the commonly used ID loss and triplet loss~\cite{luo2019trick}. This approach has demonstrated impressive performance in supervised Re-ID, seeming like highlighting the potential of prompt learning in adapting CLIP to the Re-ID task.

However, upon investigating the mechanisms that enable effective textual prompt learning in CLIP-ReID, we have made the following speculations: 1) Unlike CoOP~\cite{coop} that utilizes fixed class names to transfer textual semantics, the prompt learning in CLIP-ReID essentially learns a textual feature centroid for each ID, as shown in Figure~\ref{fig:index}. 2) The image-to-text cross-entropy loss introduced in CLIP-ReID serves as a centroid-based loss~\cite{Wen2016,Attias2017proxy}, attracting images of the same ID towards their respective textual centroids. Inspired by these findings, we propose to utilize a prototypical contrastive learning (PCL) loss~\cite{Li2021} to directly fine-tune the CLIP's image encoder without the need of prompt learning. Our PCL loss leverages the ID centroids of up-to-date visual features for fine-tuning, preventing from the disturbance caused by outdated textual centroids and potential misalignment between textual and visual features. Experimental results demonstrate that simply fine-tuning CLIP with a single PCL loss performs competitively with CLIP-ReID~\cite{clipreid}, achieving significantly higher performance compared to fine-tuning CLIP with ID loss and triplet loss.

The aforementioned fine-tuning of CLIP is performed under full supervision, where ID labels are available. In this work, we also aim to adapt CLIP to unsupervised object Re-ID. The dominant approach for unsupervised Re-ID is clustering-based, which generates pseudo labels through clustering and then use the labels to learn a Re-ID model iteratively. The iterative learning scheme and the varying number of pseudo labels at different iterations make it challenging to learn prompts similar to CLIP-ReID~\cite{clipreid}. 

Fortunately, our PCL loss-based CLIP fine-tuning fits well within this framework. Considering that state-of-the-art (SoTA) methods such as ClusterContrast~\cite{clustercontrast}, CAP~\cite{cap}, and O2CAP~\cite{O2CAP} already adopt PCL-wise losses, we can simply replace their backbone with the image encoder of CLIP and fine-tune the models using their own PCL-wise losses. However, directly replacing and fine-tuning the CLIP image encoder leads to a divergence issue due to the instability of training vision transformers under unsupervised setting. To address this issue, we adopt the patch projection layer frozen trick~\cite{Chen2021ViT-ssl}, which allows our model to converge effectively and achieve significantly higher performance compared to SoTA methods.

In summary, our work makes the following contributions:
\begin{itemize}
	\item By investigating the mechanisms of prompt learning in CLIP-ReID, we propose a prototypical contrastive learning loss (PCL)-based CLIP fine-tuning method. Our approach is simple yet effective in adapting CLIP to object Re-ID tasks. 
	\item We employ the PCL-based fine-tuning approach to adapt CLIP for various Re-ID settings, including fully supervised and unsupervised Re-ID tasks. Extensive experiments show that our fine-tuning method achieves competitive performance in both of these settings. 
\end{itemize}

\section{Related Work}
\subsection{Object Re-identification}
In this work, we investigate the adaptation of CLIP to the Re-ID task under different settings, including fully supervised and purely unsupervised Re-ID. Therefore, we provide a concise review of these two tasks.

In the past decade, fully supervised object Re-ID has witnessed remarkable advancements, primarily driven by the utilization of deep convolutional neural networks (CNNs) like ResNet-50~\cite{he2016deep}, in combination with the use of ID loss and triplet loss~\cite{hermans2017defense,luo2019trick}. Numerous CNN-based methods have been developed, leveraging multi-granularity features~\cite{sun2018beyond}, human semantics~\cite{zhang2019dsa}, attention mechanisms~\cite{chen2019abd}, etc. More recently, transformer architectures~\cite{transreid,AAformer,LA-Transformer} and pre-trained vision-language models~\cite{clip,clipreid} have also been employed to boost Re-ID performance.

Purely unsupervised Re-ID has also witnessed significant progress in recent years. A majority of research is clustering-based, which involves generating pseudo labels through clustering and iteratively learning a Re-ID model using these labels. Previous methods have made efforts to refine noisy pseudo labels~\cite{RLCC,PPLR,O2CAP}, leverage contrastive learning losses~\cite{cap,O2CAP,clustercontrast}, and design network architectures~\cite{transreidssl,TMGF}. To the best of our knowledge, the utilization of pre-trained vision-language models has not been explored in unsupervised Re-ID.  


\subsection{CLIP and Prompt Learning}
Contrastive language-image pre-training (CLIP)~\cite{clip} is a vision-language model that undergoes pre-training on millions of text-image pairs available on the Internet. It jointly trains a text encoder and an image encoder using two directional InfoNCE losses, which are extensively used in contrastive learning~\cite{Oord2019CPC,Chen2020SimCLR,he2020momentum}. Beneficial from the knowledge learned from the large-scale dataset, CLIP has demonstrated remarkable generalization ability and has been effectively transferred to various downstream vision tasks, including zero-shot recognition~\cite{coop,cocoop}, object detection~\cite{Zhong2022RegionCLIP,Wu2023CORA}, semantic segmentation~\cite{Luo2023SegCLIP}, among others. 

Prompt learning~\cite{coop,cocoop,Jia2022,Khattak2023} has gained popularity in effectively transferring CLIP to specific downstream tasks. As a pioneering work, CoOp~\cite{coop} introduces the concept of learning text prompts, which significantly outperformed hand-crafted prompts in CLIP adaptation. Building upon this, CoCoOp~\cite{cocoop} further enhances generalizability by allowing text prompts to be conditioned on inputs, VPT~\cite{Jia2022} and MaPLe~\cite{Khattak2023} extend the learning to visual and multi-modal prompts. In both textual~\cite{coop,cocoop} and multi-modal~\cite{Khattak2023} prompt learning, the class names of ground-truth labels are utilized to form input text descriptions, contributing remarkable generalization ability to unseen data~\cite{coop} and enhancing robustness to noise~\cite{Wu2023}.

However, in object re-identification, class names do not exist as the labels are represented by ID indexes. In order to adapt CLIP to Re-ID tasks, CLIP-ReID~\cite{clipreid} introduces a two-stage strategy that first learns a set of textual prompts for each ID and then employs a prompt-based loss to fine-tune CLIP, resulting in impressive performance. Nevertheless, when using our prototypical contrastive learning loss to fine-tune the CLIP image encoder, this prompt learning strategy brings almost no improvement for both fully supervised and unsupervised Re-ID tasks. These results suggest that prompt learning may not be necessary for achieving strong performance in Re-ID tasks and our PCL loss-based fine-tuning approach can be a promising alternative. 

\subsection{Prototypical Contrastive Learning}
Prototypical contrastive learning (PCL)~\cite{Li2021} is a type of contrastive learning method that operates at the cluster level. In this approach, each class or cluster is represented by a prototype, which is a central representation of the instances within that cluster. The primary objective of PCL is to attract each instance towards its own cluster's prototype while simultaneously pushing it away from the prototypes of other clusters. By leveraging the cluster-level information, prototypical contrastive learning can effectively capture local semantic structures, resulting in more effective learning compared to conventional instance-level contrastive learning methods commonly used in uni-modal and multi-modal unsupervised representation learning tasks\cite{Chen2020SimCLR,he2020momentum,clip}. In recent years, prototypical contrastive learning has been successfully applied to unsupervised Re-ID tasks. Various PCL-wise losses, such as leveraging camera-aware prototypes~\cite{cap} or online associated prototypes~\cite{O2CAP}, have been developed, leading to significant improvements in performance. In this work, we explore the use of the PCL loss for fine-tuning CLIP in both fully supervised and unsupervised Re-ID tasks.

\begin{figure*}[t]
	\centering
	\includegraphics[width=0.875\linewidth]{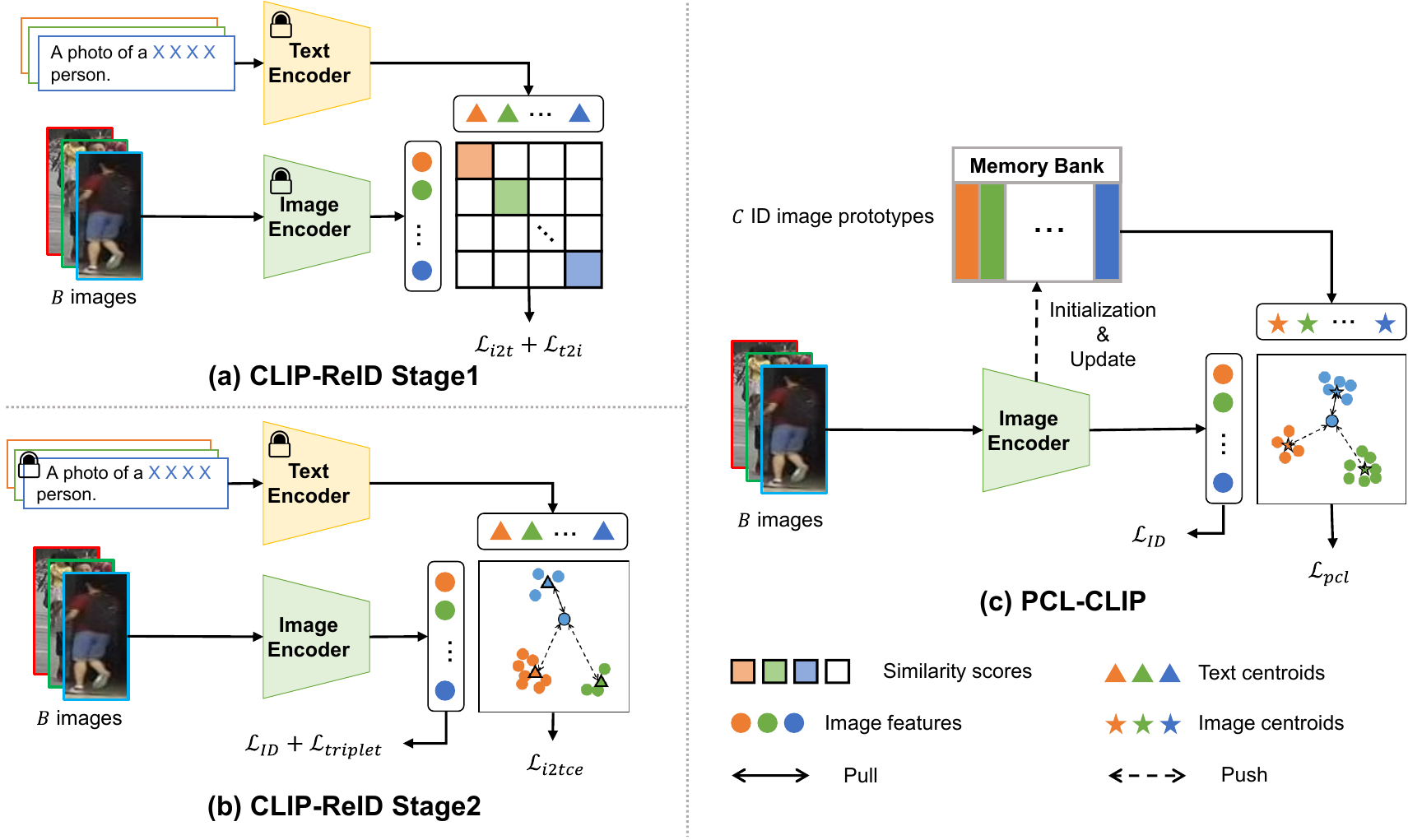}
	\caption{The framework of our PCL-CLIP model for supervised Re-ID. Different from CLIP-ReID that consists of a prompt learning stage and a fine-tuning stage, our approach directly fine-tune CLIP with a single prototypical contrastive learning (PCL) loss. In our framework, a memory bank is built to store up-to-date visual feature centroid of each ID.}
	\label{fig:framework}
	\vspace{-15pt}
\end{figure*}

\section{Revisit CLIP-ReID}
We first revisit the CLIP-ReID method~\cite{clipreid}. It proposes a two-stage strategy to adapt CLIP~\cite{clip} to supervised object Re-ID. Let us denote the pre-trained text and image encoders of CLIP as $\mathcal{T}(\cdot)$ and $\mathcal{I}(\cdot)$, respectively. These two encoders map a sequence of text prompt tokens and an image into features $\mathbf{f}^T$ and $\mathbf{f}^I$ in a joint embedding space.
%
%

In the first stage, CLIP-ReID learns a set of ID-specific text tokens while freezing both the text and image encoders like CoOP~\cite{coop}. For an image $i$ with an ID label $y_i = c$ ($c\in\{1,\cdots,C\}$), the paired text sequence input into $\mathcal{T}(\cdot)$ is designed as ``A photo of $[X]^c_1[X]^c_2\cdots[X]^c_M$ person/vehicle''. Here, $[X]^c_m$ ($m\in\{1,\cdots,M\}$) is a learnable token with the same dimension as word embedding, $C$ denotes the total number of IDs and $M$ is the nubmer of learnable tokens. The text tokens are learned by optimizing the sum of the following image-to-text and text-to-image losses:
\begin{equation}
	\label{eqn:l_i2t}
	\mathcal{L}_{i2t} = -\frac{1}{|K_i|} \sum_{k\in K_i} \log \frac{\exp (s(\mathbf{f}_k^{T}, \mathbf{f}_i^{I}) / \tau) }{\sum_{j=1}^B \exp (s(\mathbf{f}_j^{T}, \mathbf{f}_i^{I}) / \tau)}
\end{equation}
\begin{equation}
	\label{eqn:l_t2i}
	\mathcal{L}_{t2i} = -\frac{1}{|K_i|} \sum_{k\in K_i} \log \frac{\exp (s(\mathbf{f}_i^{T}, \mathbf{f}_k^{I}) / \tau) }{\sum_{j=1}^B \exp (s(\mathbf{f}_i^{T}, \mathbf{f}_j^{I}) / \tau)}
\end{equation}
where $s(\cdot,\cdot)$ represents the cosine similarity, $B$ is the batch size and $\tau$ is a temperature factor learned by CLIP. $K_i = \{ k | y_k = y_i, k\in \{1,2,...,B\} \}$ denotes the index set of positive image samples and $|\cdot|$ is the cardinality.
%

In the second stage, CLIP-ReID fine-tunes the image encoder of CLIP while fixing both the learned prompts and the text encoder. The image encoder is optimized with respect to a ID loss $\mathcal{L}_{id}$ and a triplet loss $\mathcal{L}_{tri}$ that are widely used in supervised Re-ID~\cite{luo2019trick}, together with an image-to-text cross-entropy loss defined as below:
\begin{equation}
	\label{eqn:l_i2tce}
	\mathcal{L}_{i2tce} = -\sum_{k=1}^C q_k \log \frac{\exp (s(\mathbf{f}_k^{T}, \mathbf{f}_i^{I})/ \tau) }{\sum_{j=1}^C \exp (s(\mathbf{f}_j^{T}, \mathbf{f}_i^{I}) / \tau)},
\end{equation}
where $q_k$ is a smoothed ID label. 

\textbf{Discussion.} In the CLIP-ReID framework, each ID class $c$ is associated with a textual feature $\mathbf{f}_c^{T}$, which is obtained by feeding a sequence containing the learned ID-specific prompts into the text encoder. The aforementioned loss $\mathcal{L}{i2tce}$ serves to attract the visual feature of an image towards its corresponding textual feature while pushing it away from textual features of other classes. This implies that each class's textual feature is well-aligned with its visual feature centroid. However, achieving such alignment is not guaranteed due to the following reasons. Firstly, although the training of CLIP encourages the alignment of textual and visual features in a joint embedding space, there may still be slight separation due to the modality gap~\cite{Liang2022,Shi2023}. Secondly, even if the two encoders are initially well-aligned in CLIP, fine-tuning the image encoder while keeping the text encoder fixed may introduce misalignment issues. Therefore, while the loss $\mathcal{L}{i2tce}$ significantly improves performance compared to using only the ID loss and triplet loss for fine-tuning CLIP, the presence of misalignment issues can have a negative impact on the fine-tuning process, resulting in sub-optimal performance.

\section{PCL-based CLIP Fine-tuning}
Based on our understanding of the mechanisms behind the effectiveness and potential limitations of prompt learning in CLIP-ReID~\cite{clipreid}, we propose a direct fine-tuning approach for CLIP using prototypical contrastive learning (PCL). In this work, we first apply PCL-based CLIP fine-tuning for the supervised Re-ID task and then extend our approach to unsupervised settings. By leveraging the benefits of PCL, we aim to improve the performance of CLIP in adapting to Re-ID tasks while bypassing the need for prompt learning.

\subsection{Fine-tuning for Supervised Re-ID}
\textbf{Prototypical contrastive learning.} 
Prototypical contrastive learning has been widely used in unsupervised representation learning~\cite{Li2021} and unsupervised Re-ID~\cite{cap,O2CAP,clustercontrast} tasks. The objective of PCL is to bring an instance closer to its cluster centroid while pushing it away from the centroids of other clusters. By this means, PCL learns to distinguish between different clusters and capture the underlying structure of the data in an unsupervised manner.

When applying prototypical contrastive learning to supervised Re-ID, the clusters obtained through unsupervised clustering are replaced with the ground-truth ID classes. Consequently, for an image $i$ with its corresponding visual feature $\mathbf{f}_i^I$ obtained from the image encoder of CLIP, and its ID label $y_i$, the PCL loss is defined as follows: 
\begin{equation}
	\label{eqn:l_pcl}
	\mathcal{L}_{pcl} = - \log \frac{\exp (s(\mathcal{K}[y_i], \mathbf{f}_i^{I})/ \tau) }{\sum_{j=1}^C \exp (s(\mathcal{K}[j], \mathbf{f}_i^{I}) / \tau)},
\end{equation}
in which $\mathcal{K}[j]$ represents the visual feature centroid of class $j$ stored in a memory bank $\mathcal{K}$, and the remaining symbols have the same definitions as previously mentioned.


\textbf{Memory bank.} In PCL, an external memory bank $\mathcal{K}\in R^{d\times C}$ is constructed to store the feature centroids of all ID classes. Each centroid is initialized by averaging the visual features of all images belonging to that class. During the fine-tuning the CLIP image encoder, the centroid is updated using momentum as follows:
\begin{equation}
	\mathcal{K}[y_i] \leftarrow \mu \mathcal{K}[y_i] + (1 - \mu) \mathbf{f}^{I}_i
\end{equation}
where $\mu$ is a momentum factor.

\textbf{Discussion.} In contrast to CLIP-ReID~\cite{clipreid} that utilizes a loss to attract or repel an image towards its textual feature centroid, our PCL loss operates directly on the visual features, eliminating the need for alignment between textual and visual features.  Experimental results show that fine-tuning CLIP with a single PCL loss achieves competitive performance compared to CLIP-ReID, which employs three losses for fine-tuning. Furthermore, while prototypical contrastive learning has also been successfully applied to supervised image classification~\cite{Khosla2020}, the ID loss and triplet loss remain dominant in supervised Re-ID methods. Our study indicates that the PCL loss is more effective than these two losses for adapting CLIP to supervised Re-ID tasks. 

%
%

%

\subsection{Fine-tuning for Unsupervised Re-ID}

\textbf{PCL in unsupervised Re-ID.} 
As mentioned earlier, prototypical contrastive learning has been successfully applied to unsupervised Re-ID tasks. Recent methods such as ClusterContrast~\cite{clustercontrast}, CAP~\cite{cap}, and O2CAP~\cite{O2CAP} have designed various PCL variant losses within a clustering-based framework, leading to impressive performance. For instance, ClusterContrast~\cite{clustercontrast} utilizes the PCL loss defined in Eq.(\ref{eqn:l_pcl}) based on pseudo labels generated through clustering. CAP~\cite{cap} additionally introduces an intra-camera PCL loss that divides each cluster into sub-clusters based on camera views. O2CAP~\cite{cap} incorporates an online PCL loss that rectifies noisy clusters through online association. 

Considering the similar mechanisms of prototypical contrastive learning employed in these unsupervised methods, we directly replace their feature extraction backbones with the CLIP image encoder and utilize their respective losses for fine-tuning. By leveraging these established techniques, we aim to leverage the strengths of PCL in unsupervised Re-ID tasks while benefiting from the powerful feature extraction capabilities of the CLIP image encoder.

\textbf{Divergence issue.}
However, directly fune-tuning the vision transformer (ViT)-based image encoder of CLIP for unsupervised Re-ID leads to divergence. This divergence issue is similar to the instability observed by Chen et al.\cite{Chen2021ViT-ssl} during their self-supervised learning of vision transformers. To mitigate this problem, we adopt the trick proposed by them~\cite{Chen2021ViT-ssl} and freeze the patch projection layer. The parameters of this layer remain unchanged throughout the entire fine-tuning process, as they were pre-trained in CLIP. By employing this strategy, the fine-tuning process is stabilized and better converged.

%

\textbf{Discussion.} 
The clustering-based framework for unsupervised Re-ID involves conducting clustering and learning Re-ID models iteratively. This iterative process poses challenges when employing CLIP-ReID~\cite{clipreid} to adapt CLIP, as the number of pseudo ID labels varies at each iteration, making it cumbersome to perform prompt learning. In contrast, our single-stage approach fine-tunes CLIP without prompt learning, providing a more convenient solution for adapting CLIP to unsupervised Re-ID tasks.

%
%
%

%

\section{Experiments}

\subsection{Datasets \& Evaluation Metrics}
We evaluate the proposed method on two person Re-ID datasets: Market-1501~\cite{7410490} and MSMT17~\cite{wei2018person}, as well as one vehicle Re-ID dataset: VeRi-776~\cite{liu2018veri}. Consistent with common practices, we utilize the mean Average Precision (mAP) and Cumulative Matching Characteristic (CMC) at Rank-1, Rank-5, and Rank-10 as the evaluation metrics. 



\subsection{Implementation Details}
We utilize the ViT-based image encoder of CLIP~\cite{clip} for fine-tuning. Specifically, we select the ViT-B/16 backbone, which consists of $12$ transformer layers, with each layer employing $6$ attention heads. Following CLIP-ReID~\cite{clipreid}, we pass the output of the encoder through a linear projection layer to reduce the feature dimension from $768$ to $512$. Additionally, we incorporate BNNeck~\cite{luo2019trick} to both the output layer and the linear projection layer. The features generated from the two BNNecks are concatenated and L2-normalized to produce the final visual feature of an image. 

In addition, each input image is resized to $256\times 128$ and augmented with random horizontal flipping, cropping and random erasing~\cite{random-erase}. For the fine-tuning process, we utilize the SGD optimizer with the learning rate of $3.5\times10^{-4}$ and the weight decay of $5\times 10^{-4}$. In supervised Re-ID, we fine-tune the PCL-based CLIP model for $50$ epochs, with $200$ iterations per epoch. Other settings are kept the same as CLIP-ReID. When applying our PCL-based fine-tuning technique to unsupervised Re-ID methods, such as ClusterContrast~\cite{clustercontrast}, CAP~\cite{cap}, and O2CAP~\cite{O2CAP}, we maintain most of the original settings from these methods. Moreover, all experiments are conducted on a single RTX A6000 GPU using the PyTorch toolkit~\cite{pytorch}.

%
%
%
%
%
%

\subsection{Ablation Studies}
We first conduct a series of experiments to validate the effectiveness of our proposed method. These experiments are carried out on the MSMT17 dataset.

\textbf{Is prompt learning necessary for adapting CLIP to supervised Re-ID?}
We begin by investigating the necessity of prompt learning when adapting CLIP to supervised Re-ID. In Table~\ref{tab:my-table}, from the results of Baseline2 vs. CLIP-ReID3, we observe that when fine-tuning CLIP with the ID loss $\mathcal{L}_{id}$ and the triplet loss $\mathcal{L}_{tri}$, prompt learning (\ie, using the loss $\mathcal{L}_{i2tce}$ that is based on learned prompts) significantly improves the model's performance, as demonstrated in the work of CLIP-ReID~\cite{clipreid}. However, when the prototypical contrastive learning loss $\mathcal{L}_{pcl}$ is employed for CLIP fine-tuning, the inclusion of the loss $\mathcal{L}_{i2tce}$ actually hampers the performance, as shown by the results of PCL-CLIP1 vs. PCL-CLIP2. This suggests that prompt learning is not necessary when an appropriate training loss is chosen.

\begin{figure}[h]
	\centering
	\includegraphics[width=\linewidth]{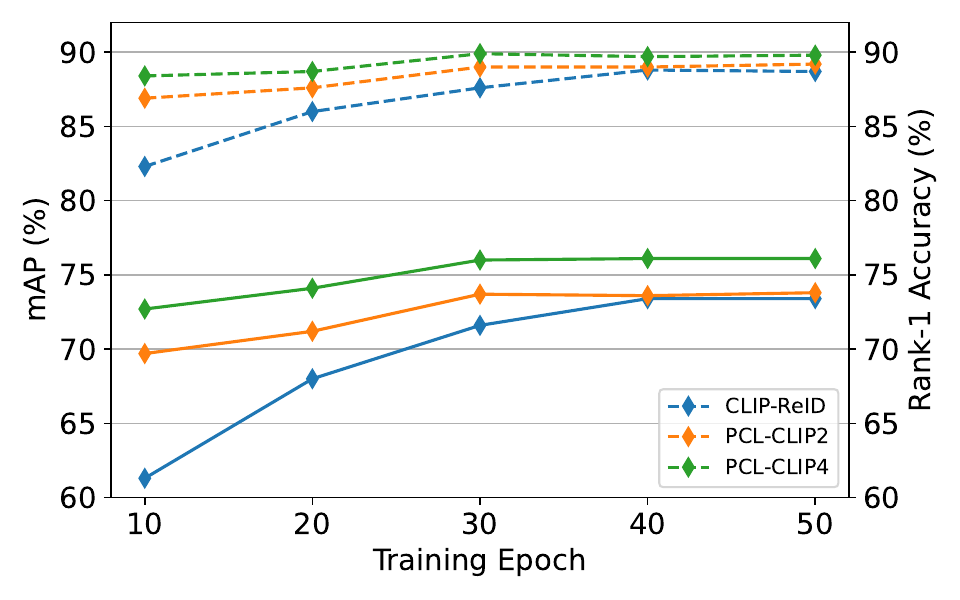}
	\caption{The performance of CLIP-ReID, PCL-CLIP2, and PCL-CLIP4 varying during the fine-tuning process. The solid line denotes mean average precision (mAP) and the dash line denotes the rank-1 accuracy. Best view with color.}
	\label{fig:mAP}
	\vspace{-15pt}
\end{figure}

\textbf{How does different losses affect the fine-tuning of CLIP?} 
The aforementioned experiments highlight the critical role of the loss function in shaping the learned representations and influencing the performance of the fine-tuned CLIP model. To further explore this, we conduct experiments using different loss functions for fine-tuning and report the results in Table~\ref{tab:my-table}. As demonstrated by Baseline1-2 and PCL-CLIP2-5, fine-tuning with the ID loss alone yields inferior performance. However, utilizing a single PCL loss allows us to achieve highly competitive results. Furthermore, the combination of the PCL loss with the ID loss further enhances the performance. Figure~\ref{fig:mAP} depicts the mAP and Rank-1 performance of PCL-CLIP2, PCL-CLIP4, and CLIP-ReID as they vary across iterations. It demonstrates that both PCL-CLIP2 and PCL-CLIP4 also converge at a faster rate.


\textbf{Is prompt learning necessary for adapting CLIP to unsupervised Re-ID?}
We further conduct an experiment to investigate whether prompt learning can improve performance in unsupervised scenarios. However, due to the iterative learning mechanism and the varying number of clusters at different epochs, it is not straightforward or cumbersome to adopt the prompt learning strategy of CLIP-ReID for unsupervised Re-ID. Therefore, we conduct two experiments: one with prompt learning applied every 10 epochs and another with prompt learning applied only at the last epoch. The results are presented in Table~\ref{tab:my-table2}. The results indicate that prompt learning, even when applied every 10 epochs, only leads to a marginal improvement in performance compared to our PCL-based direct fine-tuning approach. This suggests that in unsupervised scenarios, the benefits of prompt learning may not be significant also.

\begin{table}[]
	\resizebox{\linewidth}{!}{
		\begin{tabular}{l|cccc|cc}
			\hline
			\multirow{2}{*}{Methods}     & \multicolumn{4}{c|}{Loss Function}                                    & \multicolumn{2}{c}{MSMT17} \\ \cline{2-7} 
			& $\mathcal{L}_{id}$ & $\mathcal{L}_{tri}$ & $\mathcal{L}_{i2tce}$ & $\mathcal{L}_{pcl}$ & mAP        & Rank-1        \\ \hline
			Baseline1 & \checkmark & & & & 44.5 & 70.0 \\ 
			Baseline2 & \checkmark  & \checkmark       &              &              & 66.2          & 84.3             \\
			\hline
			CLIP-ReID1 &			   & 				  &	\checkmark	 &				&	49.5			&	76.8			   \\
			CLIP-ReID2 &             & \checkmark       & \checkmark   &              & 71.4          & 87.8             \\
			CLIP-ReID3 & \checkmark  & \checkmark       & \checkmark   &              & 73.4         & 88.7             \\
			\hline
			PCL-CLIP1                      &             &                  &     \checkmark         & \checkmark   & 71.2  & 87.4  \\ 
			\hline
			\rowcolor{mygray}
			PCL-CLIP2                      &             &                  &              & \checkmark   & 73.8          & 89.2             \\
			PCL-CLIP3                      &             & \checkmark       &             & \checkmark   & 73.9          & 88.7             \\
			\rowcolor{mygray}
			PCL-CLIP4                      & \checkmark   &        &             & \checkmark   &     76.1       &      89.8         \\
			PCL-CLIP5                     & \checkmark   & \checkmark       &             & \checkmark   & 76.1          & 89.6             \\
			\hline
		\end{tabular}
	}
	\caption{Ablation study on loss functions used for fine-tuning CLIP for supervised Re-ID.}
	\label{tab:my-table}
\end{table}

\begin{table}[]
	\resizebox{\linewidth}{!}{
		\begin{tabular}{l|c|cc|cc}
			\hline
			\multirow{2}{*}{Methods}   & \multirow{2}{*}{Prompt Learning}  & \multicolumn{2}{c|}{Loss Function}  & \multicolumn{2}{c}{MSMT17} \\ \cline{3-6} 
			&     & $\mathcal{L}_{i2tce}$ & $\mathcal{L}_{pcl}$ & mAP        & Rank-1        \\ \hline
			PCL-CLIP$_{\text{O2CAP}}$1                &  every 10 ep   &     \checkmark         & \checkmark   & 66.2  & 85.8  \\ 
			PCL-CLIP$_{\text{O2CAP}}$2                 &  last ep   &     \checkmark         & \checkmark   & 65.7  & 85.0 \\ \hline
			PCL-CLIP$_{\text{O2CAP}}$3                 &  None   &      & \checkmark   & 65.5          & 84.9             \\
			\hline
		\end{tabular}
	}
	\caption{Ablation study on the use of prompt learning for adapting CLIP to unsupervised Re-ID.}
	\label{tab:my-table2}
\end{table}

\begin{table*}[ht]
		\centering
		\resizebox{\textwidth}{!}{
			\begin{tabular}{c|l|cccc|cccc}
				\hline
				\multicolumn{1}{c|}{\multirow{2}{*}{Backbone}} 
				& \multicolumn{1}{c|}{\multirow{2}{*}{Methods}}             
				& \multicolumn{4}{c|}{Market1501}                                        
				& \multicolumn{4}{c}{MSMT17}    \\ \cline{3-10} 
				& \multicolumn{1}{c|}{}                          
				& mAP  & Rank-1  & Rank-5    &   Rank-10 & mAP  & Rank-1 & Rank-5 & Rank-10 \\ \hline
				\multicolumn{10}{l}{\textit{Fully supervised methods}} \\ \hline
				\multicolumn{1}{c|}{\multirow{5}{*}{CNN}}      
				& ABD-Net$_{\text{\color{gray}{ICCV19}}}$\text{\textdaggerdbl}~\cite{ABD-Net} & 88.3 & 95.6   & -   & -       & 60.8 & 82.3   & 90.6   & -       \\
				& OSNet$_{\text{\color{gray}{ICCV19}}}$~\cite{osnet} & 84.9 & 94.8  & -   & -       & 52.9 & 78.7   & -      & -       \\
				&  SAN$_{\text{\color{gray}{AAAI20}}}$~\cite{SAN} & 88.0 & 96.1   & -   & -       & 55.7 & 79.2   & -      & -       \\
				&  CDNet$_{\text{\color{gray}{CVPR21}}}$~\cite{CDNet} & 86.0 & 95.1 & -   & -       & 54.7 & 78.9   & -      & -       \\
				&  CAL$_{\text{\color{gray}{ICCV21}}}$\text{*}~\cite{CAL} & 89.5 & 95.5   & -   & -       & 64.0 & 84.2   & -      & -       \\
				\hline 
				\multicolumn{1}{c|}{\multirow{5}{*}{ViT}} & AAformer$_\text{\color{gray}{arXiv21}}$\text{\textdagger}~\cite{AAformer} & 87.7 & 95.4 & - & - & 63.2 & 83.6 & - & - \\
				\multicolumn{1}{c|}{}                       
				& TransReID$_{\text{\color{gray}{ICCV21}}}$~\cite{transreid} &  89.5 & 95.2     & -   & -       & 69.4 & 86.2   & -      & -       \\
				&  CLIP-ReID$_{\text{\color{gray}{AAAA23}}}$~\cite{clipreid}  & 89.6 & 95.5  & -  & -   & 73.4 & 88.7   & -      & -       \\
				\rowcolor{mygray}
				&  PCL-CLIP$_{\mathcal{L}_{pcl}}$    & 91.0 & 96.1   & 98.8   &  99.2 & 73.8 & 89.2   & 94.7   & 95.8    \\
				\rowcolor{mygray}
				&  PCL-CLIP$_{\mathcal{L}_{pcl}+\mathcal{L}_{id}}$    & 91.4 & 95.9   & 98.5   &  99.2 & 76.1 & 89.8   & 94.7   & 96.0    \\
				\hline
				\multicolumn{10}{l}{\textit{Unsupervised methods}}  \\ \hline
				\multicolumn{1}{c|}{\multirow{6}{*}{CNN}}
				&  CAP$_{\text{\color{gray}{AAAI21}}}$~\cite{cap}  & 79.2 & 91.4  & 96.3 &  97.7 & 36.9 & 67.4   & 78.0   & 81.4    \\ 
				&  CC$_{\text{\color{gray}{ACCV22}}}$~\cite{clustercontrast}  & 83.0 & 92.9   & 97.2 & 98.0  & 31.2 & 61.5   & 71.8   & 76.7    \\                
				&  ICE$_{\text{\color{gray}{ICCV21}}}$~\cite{ICE} & 82.3 & 93.8   & 97.6  & 98.4  & 38.9 & 70.2   & 80.5   & 84.4 \\
				& O2CAP$_{\text{\color{gray}{TIP22}}}$~\cite{O2CAP} & 82.7 & 92.5 & 96.9 &  98.0 & 42.4 & 72.0   & 81.9   & 85.4    \\
				& PPLR$_{\text{\color{gray}{CVPR22}}}$\text{\textdaggerdbl}~\cite{PPLR}  & 84.4 & 94.3  & 97.8  &  98.6 & 42.2 & 73.3   & 83.5   & 86.5    \\
				&  DiDAL$_{\text{\color{gray}{TMM23}}}$~\cite{liu2023discriminative} & 84.8 & 94.2  & 98.2 & -  & 45.4 & 74.0   & 84.3   & -       \\ \hline
                \multicolumn{1}{c|}{\multirow{3}{*}{ViT}}
				&  TransReID-SSL$_{\text{\color{gray}{arXiv21}}}$~\cite{transreidssl} & 89.6 & 95.3   & -  &  - & 50.6 & 75.0   & -      & -       \\
				& TMGF$_{\text{\color{gray}{WACV23}}}$\text{\textdaggerdbl}~\cite{TMGF}  & 89.5 & 95.5  & 98.0  &  98.7 & 58.2 & 83.3   & 90.2   & 92.1    \\
				\rowcolor{mygray}
				&  PCL-CLIP$_{\text{CC}}$  & 86.9 &  94.2   & 97.8 & 98.7  & 56.4  &  77.9   &  85.2 & 87.2   \\
				\rowcolor{mygray}
				&  PCL-CLIP$_{\text{CAP}}$  & 87.4  &  93.9   & 97.7 & 98.5  & 53.6  & 79.0  & 88.4  &  91.1   \\
				\rowcolor{mygray}
				&  PCL-CLIP$_{\text{O2CAP}}$  & 88.4 & 94.8    & 98.0   &  98.7 & 65.5 & 84.9   & 92.0   & 94.0    \\
				\hline
			\end{tabular}
		}
		\caption{Comparison with state-of-the-art methods on person Re-ID datasets. The symbol \text{\textdagger},  \text{\textdaggerdbl} and \text{*} denote that the input images for the respective methods are of size $384\times256$,  $384\times128$ and $384\times192$, while all other methods utilize images of size $256\times 128$.}
		\label{tab:sota}
	\end{table*}

	\begin{table}[]
		\centering
		\resizebox{\linewidth}{!}{
			\begin{tabular}{l|cccc}
				\hline
				\multicolumn{1}{c|}{\multirow{2}{*}{Method}} & 
				\multicolumn{4}{c}{VeRi-776} \\ \cline{2-5} 
				& mAP & Rank-1 & Rank-5 & Rank-10 \\ \hline
				\multicolumn{5}{l}{\textit{Fully supervised methods}}  \\ \hline
				VehicleNet$_{\text{\color{gray}{arXiv20}}}$~\cite{VehicleNet}  & 83.4 & 96.8 & - &  -       \\
				SAN$_{\text{\color{gray}{AAAI20}}}$\cite{SAN}  & 72.5 & 93.3 & - & - \\
				CAL$_{\text{\color{gray}{ICCV21}}}$\cite{CAL} & 74.3 & 95.4 & 97.9 &  -       \\
				TransReID$_{\text{\color{gray}{ICCV'21}}}$\cite{transreid}  & 82.0 & 97.1 & - &  -       \\
				CLIP-ReID$_{\text{\color{gray}{AAAI'23}}}$\cite{clipreid}  & 83.3 & 97.4 & - &  -       \\
				\rowcolor{mygray}
				PCL-CLIP$_{\mathcal{L}_{pcl}}$       & 81.1 & 97.0 & 98.7 & 99.3       \\
				\rowcolor{mygray}
				PCL-CLIP$_{\mathcal{L}_{pcl}+\mathcal{L}_{id}}$ & 82.5 & 97.1 & 98.6 & 99.2 \\
				\hline
				\multicolumn{5}{l}{\textit{Unsupervised methods}}                                     \\ \hline
				MMT$_{\text{\color{gray}{arXiv21}}}$~\cite{MMT} &  35.3   &    74.6    &    82.6    &  87.0 \\
				SpCL$_{\text{\color{gray}{NeurIPS20}}}$~\cite{SpCL}  &  36.9   &    79.9    &    86.8    &  89.9 \\
				CC$_{\text{\color{gray}{ACCV22}}}$~\cite{clustercontrast}  &  40.3   &    84.6    &    89.2    &  91.6 \\
				CAP$_\text{\color{gray}{AAAI21}}$~\cite{cap} & 41.0 & 86.7 & 90.5 & 92.9 \\
				O2CAP$_{\text{\color{gray}{TIP22}}}$~\cite{O2CAP} & 41.9 & 87.5 & - & - \\
				RLCC$_{\text{\color{gray}{CVPR21}}}$~\cite{RLCC}  & 39.6 & 83.4 & 88.8 & 90.9 \\
				PPLR$_{\text{\color{gray}{CVPR22}}}$~\cite{PPLR}  & 41.6 & 85.6 & 91.1 & 93.4 \\
				\rowcolor{mygray}
				PCL-CLIP$_{\text{CC}}$   & 34.7 & 72.7  & 78.6 & 82.9  \\
				\rowcolor{mygray}
				PCL-CLIP$_{\text{CAP}}$   & 44.2 & 89.9 & 93.4 & 94.8  \\
				\rowcolor{mygray}
				PCL-CLIP$_{\text{O2CAP}}$   & 45.5 & 90.7 & 93.9 & 95.0 \\
				\hline
			\end{tabular}
		}
		\caption{Comparison with state-of-the-art methods on vehicle Re-ID dataset VeRi-776.}
		\label{tab:veri-sota}
	\end{table}
	
	\subsection{Comparison to State-of-the-Art}
	Finally, we evaluate the performance of our proposed method, referred to as PCL-CLIP, against state-of-the-art methods on person and vehicle Re-ID datasets. The comparative results are presented in Table~\ref{tab:sota} and Table~\ref{tab:veri-sota}, respectively.
	
	\textbf{Comparison on supervised person Re-ID.} 
	In the supervised person Re-ID task, we compare our approach with eight recent methods. Among these methods, five~\cite{ABD-Net,osnet,SAN,CDNet,CAL} utilize CNN-based backbones, while three~\cite{AAformer,transreid,clipreid} employ ViT-based backbones. Notably, both CLIP-ReID~\cite{clipreid} and our approach adapt the pre-trained CLIP model, while the  remaining methods are pre-trained on ImageNet. It is evident that the use of large-scale pre-trained models yields significant performance improvements compared to the other methods, particularly on the MSMT17 dataset. Our method achieves competitive performance with CLIP-ReID when a single PCL loss is employed. However, when incorporating an ID loss, our method exhibits a substantial improvement on MSMT17.
	
	\textbf{Comparison on unsupervised person Re-ID.}
	In unsupervised person Re-ID, we compare our approach with eight recent methods as well. Among these methods, six~\cite{cap,clustercontrast,ICE,O2CAP,PPLR,liu2023discriminative} utilize CNN-based backbones and two~\cite{transreidssl,TMGF} employ ViT-based backbones. To evaluate the performance of our CLIP fine-tuning approach, we apply it to ClusterContrast (CC)~\cite{clustercontrast}, CAP~\cite{cap}, and O2CAP~\cite{O2CAP} by replacing their original backbones with the CLIP image encoder. Comparing our CLIP fine-tuning approach with these methods~\cite{clustercontrast, cap, O2CAP}, we observe significant performance improvements on both the Market1501 and MSMT17 datasets. Additionally, TransReID-SSL~\cite{transreidssl} and TMGF~\cite{TMGF} utilize ViT backbones that are pre-trained on the large-scale unlabeled dataset LUPerson~\cite{LUPerson}. However, our CLIP fine-tuning approach with O2CAP, outperforms them by a considerable margin on the MSMT17 dataset. These results demonstrate the effectiveness of the CLIP fine-tuning approach in unsupervised scenarios, surpassing existing methods and achieving notable performance improvements on challenging datasets.


	\textbf{Comparison on vehicle Re-ID.}
	Table~\ref{tab:veri-sota} presents the comparison results on a vehicle Re-ID dataset. The table includes five fully supervised methods~\cite{VehicleNet,SAN,CAL,transreid,clipreid} and seven unsupervised methods~\cite{MMT,SpCL,clustercontrast,cap,O2CAP,RLCC,PPLR} for reference. The proposed method achieves comparable results with state-of-the-art methods in the fully supervised setting. In unsupervised scenarios, our CLIP fine-tuning approach, specifically with the O2CAP loss, outperforms previous methods by a considerable margin. 
	
	\section{Conclusion}
	In this work we have presented a simple yet effective approach to adapt CLIP to both supervised and unsupervised Re-ID tasks. Our approach involves directly fine-tuning the image encoder of CLIP using a single prototypical contrastive learning (PCL) loss, eliminating the need for prompt learning. Remarkably, our method achieves competitive performance compared to CLIP-ReID, which requires both prompt learning and fine-tuning. Moreover, by incorporating the PCL loss alongside the ID loss during fine-tuning, we observe a significant improvement over CLIP-ReID on MSMT17. Our findings highlight the potential of our approach to simplify the adaptation of CLIP for Re-ID tasks while achieving comparable or even superior performance to existing methods.

	{\small
		\bibliographystyle{ieee_fullname}
		\bibliography{egbib}
	}
	
\end{document}